\documentclass[]{interact}

\usepackage{graphicx}
\usepackage{amssymb}
\usepackage{amsmath}
\usepackage{cite}
\usepackage{comment}
\usepackage{color}
\usepackage{url}
\usepackage{alltt}
\usepackage{cleveref}
\usepackage{tikz}
\usepackage{pgfplots}
\usepackage{pgf-umlsd}
\usepackage{cprotect}
\usepackage{multirow}
\usepackage{algorithm}
\usepackage[noend]{algpseudocode}
\usepackage{bigfoot}
\usepackage{paralist}
\usepackage{tabularx}
\usepackage[whole]{bxcjkjatype}
\usepackage{ifthen}
\usepackage{threeparttable}
\usepackage[hang,small,bf]{caption}
\usepackage[subrefformat=parens]{subcaption}
\usepackage[super]{nth}
\usepackage{colortbl}
\usepackage{scalefnt}
\usepackage{threeparttable}
\usepackage[bottom]{footmisc}

\usepackage{siunitx}
\usepackage{nth}
\usepackage[labelsep=space]{caption}
\usepackage[keeplastbox]{flushend}
\usepackage{bm}

\newcounter{num}

\newcommand{\rnum}[1]{\setcounter{num}{#1} \roman{num}}

\pgfplotsset{compat=1.14}
\DeclareMathOperator*{\argmin}{argmin}

\makeatletter
\newcommand{\figcaption}[1]{\def\@captype{figure}\caption{#1}}
\newcommand{\tblcaption}[1]{\def\@captype{table}\caption{#1}}
\makeatother

\newcommand{\objFunc}[1]{\sum_{t} \|\boldsymbol{#1}_{t}\|^{2}_{2}} 

\begin{document}

\articletype{Full paper}

\title{Learning-based Collision-free Planning on Arbitrary Optimization Criteria in the Latent Space through cGANs}

\author{
    \name{
        \thanks{CONTACT: tomoki\_a@fuji.waseda.jp, \{iino, mori\}@idr.ias.sci.waseda.ac.jp, ryota.torishima@gmail.com,\\\{takahashi, guguchi, hillbig\}@preferred.jp, ogata@waseda.jp}
        \thanks{$^{a}$ Waseda University \& AIST, Tokyo, Japan}
        \thanks{$^{b}$ SoftBank Corp., Tokyo, Japan}
        \thanks{$^{c}$ Preferred Networks, Inc., Tokyo, Japan}
        Tomoki Ando\textsuperscript{a},
        Hiroto Iino\textsuperscript{a},
        Hiroki Mori\textsuperscript{a, 1}\thanks{$^{1}$ H. Mori was a researcher in Cergy-Pontoise University when he came up with the idea.},
        Ryota Torishima\textsuperscript{b, 2}\thanks{$^{2}$ This work was an achievement while he was at Waseda University \& AIST.},\\
        Kuniyuki Takahashi\textsuperscript{c},
        Shoichiro Yamaguchi\textsuperscript{c},\\
        Daisuke Okanohara\textsuperscript{c},
        and Tetsuya Ogata\textsuperscript{a}
    }
}
\maketitle

\setcounter{footnote}{2}
\begin{abstract}
We propose a new method for collision-free planning using Conditional Generative Adversarial Networks (cGANs) to transform between the robot's joint space and a latent space that captures only collision-free areas of the joint space, conditioned by an obstacle map.
Generating multiple plausible trajectories is convenient in applications such as the manipulation of a robot arm by enabling the selection of trajectories that avoids collision with the robot or surrounding environment.
In the proposed method, various trajectories that avoid obstacles can be generated by connecting the start and goal state with arbitrary line segments in this generated latent space.
Our method provides this collision-free latent space, after which \emph{any} planner, using \emph{any} optimization conditions, can be used to generate the most suitable paths on the fly.
We successfully verified this method with a simulated and actual UR5e 6-DoF robotic arm.
We confirmed that different trajectories could be generated depending on optimization conditions.
\footnote{An accompanying video is available at the following link:\\ \url{https://www.youtube.com/watch?v=IJUxdmaSwy0}}
\end{abstract}

\begin{figure}[t]
    \centering
    \includegraphics[width=0.90\columnwidth]{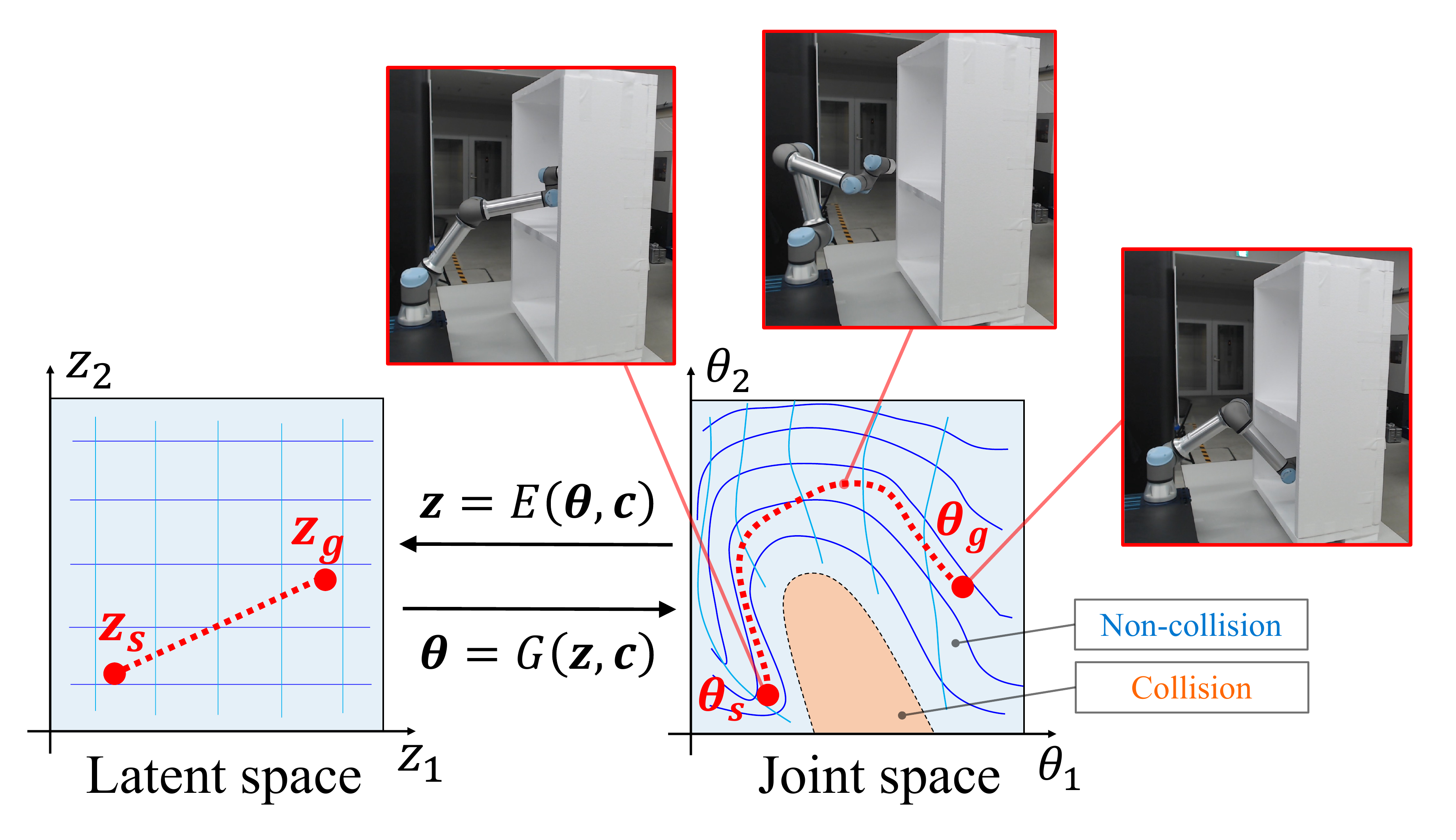}
    \caption{
		Collision-free planning for robot arm using latent space of cGANs.
		Latent variables $\boldsymbol{z}_{s}$ and $\boldsymbol{z}_{g}$ that correspond to the start $\boldsymbol{\theta}_{s}$ and goal $\boldsymbol{\theta}_{g}$ joint angles of the robot arm.
    	Any path in the latent space $\boldsymbol{z}_{s:g}$ is mapped to a collision-free path in the joint space $\boldsymbol{\theta}_{s:g}$ using generator ${G}$ with condition $\boldsymbol{c}$ as obstacle information.
    	On the other hand, $E$ is an inverse transformation of the generator~${G}$, in which the joint angles map to the latent variables.}
    \label{fig:method_transform}
\end{figure}

\section{Introduction}
\label{sec:introduction}
Collision-free planning is essential for robots working in various environments.
Multiple (potentially infinite) paths exist from a given start to a goal.
It is necessary to calculate the optimal path according to certain desired criteria, such as the minimization of the velocities, acceleration, or jerk for the robot's movements (\rnum{1}~)~\textbf{\emph{Customizability}}).
There are other two important factors in collision-free planning: \rnum{2}~)~\textbf{\emph{Adaptability}} and \rnum{3}~)~\textbf{\emph{Scalability of computation}}.
\rnum{2}~)~Robots need to adapt quickly to a new environment, which requires appropriate planning for the placement of untrained obstacles.
\rnum{3}~)~These planning operations should be calculable, even when there are many obstacles, since it generally takes a long time to collision-check for a large number of obstacles.
In other words, calculation time should scale well with the number of obstacles.

It is challenging to meet these three requirements using existing methods (see \Cref{sec:related works}).
Contrary to traditional planning in Cartesian or joint space, we propose to plan trajectories within a new collision-free space as follows.
Our method consists of two steps: 1) construction of latent space and 2) planning using this latent space.
Conditional Generative Adversarial Networks (cGANs)~\cite{goodfellow2014generative, mehdi2014conditional} are used to map joint space to latent space corresponding to its collision-free areas, such that the robot does not collide with obstacles if a path is planned within this latent space (See Fig.~\ref{fig:method_transform}).
That is, selecting any point in the latent space yields a particular robot pose that does not collide with obstacles.
There are several methods for acquiring such a latent space.
We use GANs since they offer the advantage of arbitrarily determining the distribution of the latent space.
The uniform distribution of $[0,1]$ as a latent space allows the region to be clearly defined and is a convex space.
The mappings from the latent space to joint space \textbf{\emph{adapts}} to various environments by changing according to obstacle information given to the cGANs as a condition.
The latent space is collision-free; any point on a line connecting any two points is also within this latent space (since this is a convex space).
Thus, a collision-free path can be generated by connecting the start and goal states with any arbitrary line or curve in the latent space within a domain of definition.
Then, the joint trajectory corresponding to the trajectory planned in latent space is constructed through the learned mappings.
Since we separated the learning of the mappings and the actual planning (or trajectory selecting), we can generate \textbf{any} trajectory we want on the fly for \textbf{any} optimization parameters that we want without considering collisions, making our method highly \textbf{\emph{customizable}}.
Furthermore, since planning is performed in the latent space without colliding with obstacles, ideally there is no need for collision-check for obstacles.
However, in practical use, it is challenging to guarantee 100\% obstacle avoidance using only learning-based methods. The learning method can generate a trajectory but may slightly collide with obstacles. A collision check is performed for the generated trajectory, and only the collision points are modified using the existing planning method (in our case RRT Connect~\cite{kuffner2000rrt}). The computation time for path planning can be reduced compared to existing planning methods for all trajectories.
The computation time for our proposed learning method does not depend on the number of obstacles, making it \textbf{\emph{scalable}} for complex environments.
The most significant advantage compared to existing methods is \textbf{\emph{customizability}}, where trajectories satisfying arbitrary optimization criteria can be easily generated in this latent space.
The adaptability of trajectory generation to changes in the environment and the computational time for the 6-DoF robot were also evaluated, showing the potential for future expansion.
\section{Related Work}
\label{sec:related works}
There are mainly two planning methods: model-based and learning-based.
The following two model-based methods are the most common: Design functions for obstacles and goals (e.g., potential fields~\cite{warren1990multiple, li2012efficient} and Riemannian motion policies with improved potential fields~\cite{ratliff2018riemannian}), search and optimization (e.g., RRTs~\cite{lavalle1998rapidly, kuffner2000rrt, lavalle2001randomized, wang2010triple},  $A^{\ast}$~\cite{hart1968formal}, and gradient-based approach~\cite{campana2016gradient}).
Methods that combine these are also proposed and generally show improved results~\cite{sertac2011incremental, naderi2015rtrrt, qureshi2015intelligent, ahmed2017potential, tahir2018potentially}.
While model-based methods can reliably avoid obstacles, their adaptability to various environments in real-time is limited since these methods require specific function design and adjustment of parameters for each situation in advance, not to mention the enormous computational searching cost.
As model-based methods are usually calculated according to certain conditions/criteria, such as shortest traveling distance in end-effector space or joint space~\cite{Lalibertk1994RedundantManipulator} or minimum jerk change~\cite{Flash1985MJ}, other calculations must be performed when these criteria change.
In other words, \textbf{model-based methods lack \emph{scalability} and \emph{customizability}}.

The data collected by the model-based methods can be used to train learning-based algorithms, particularly deep learning~\cite{srinivas2018universal, aviv2016value, ichter2018learning, angelina2019learning, kumar2019lego, Terasawa20203d, ota2020efficient, wen2018path, taniguchi2022planning}.
These algorithms can infer a path for a new environment in a short time if it has been trained sufficiently in advance.
However, learning-based methods have the challenge that only one or a few paths can be generated, and what kind of paths are generated depends on the training data.
For example, if naive RRT is used as training data, only collision-free paths to the goal will be generated during inference, usually without considering any additional constraints that naive RRT also does not.
Usually, \textbf{learning-based methods lack \emph{customizability}}.

In~\cite{kutsuzawa2019motion, ahmed2018motion}, the authors studied the generation of multiple trajectories.
Since the target of~\cite{kutsuzawa2019motion} was to generate various trajectories in environments with no obstacles, obstacle avoidance was out of their scope.
Another method is a learning-based approach with deep learning to perform dimensionality compression and planning in latent space to handle high-dimensional information~\cite{ichter2019robot}.
Their approach requires collision detection for each planning process.
Our proposed method is to plan paths in a \textbf{collision-free space} which are mapped from the latent space to joint space.
Hence, our method does not require a collision check for each planning process.
Since the trajectory of~\cite{ahmed2018motion} is fixed once it is generated, at best, only the optimal trajectory among the ones generated can be selected, which is not necessarily the best for the situation at hand.
Thus, they have to generate trajectories until one of them satisfies the criteria necessary for the situation, but they are generated randomly, and the method does not provide a way to define optimality.
Our method does not directly output the trajectories but provides a collision-free space after which any planner, using any optimization conditions, can generate \textbf{the most suitable paths}.

The contribution of this research is to realize optimized planning with the three factors;
\rnum{1}~)~\textbf{\emph{Customizability}}, \rnum{2}~)~\textbf{\emph{Adaptability}} , and \rnum{3}~)~\textbf{\emph{Scalability of computation}}.
\begin{figure}[t]
    \centering
    \includegraphics[width=0.95\textwidth]{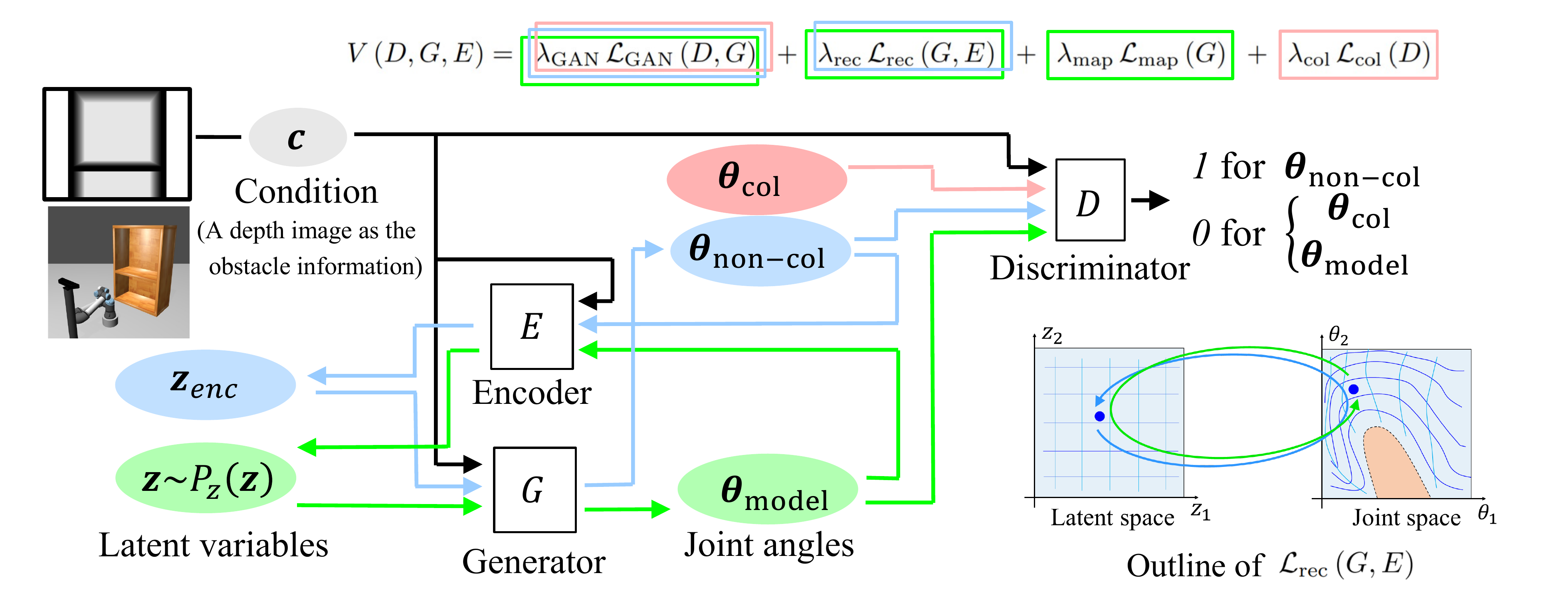}
     \caption{Structure of collision-free planning model using cGANs.
     The models are optimized with the four objective functions with coefficients~$\lambda$s of each $\mathcal{L}$.
     }
     \label{fig:network}
\end{figure}
\section{Method}
\label{sec:method}
Our proposed method consists of the following two steps: 1) Construction of a latent space corresponding to the joint space to avoid collision for \textbf{\emph{Adaptability}} (Section~\ref{sec:Training cGANs}), and 2) planning according to the objective using the constructed latent space for \textbf{\emph{Customizability}} and \textbf{\emph{Scalability of computation}} (Section~\ref{sec:path_planning}).
\subsection{Training cGANs}
\label{sec:Training cGANs}
We propose a method that maps the latent space of cGANs to the collision-free area of the robot's joint space so that the robot learns not to collide with obstacles.
Thus, any planned path in that latent space can be associated with a collision-free path in joint space.
The mapping from the latent space to the joint space \textbf{\emph{adapts}} accordingly to the obstacle information given to cGANs as a condition.
The correspondence from the latent space to joint space is trained by cGANs, which uses a min-max game between a generator $G$ and a discriminator $D$. 
Also, encoder $E$ is trained to approximate an inverse transformation of $G$; i.e., $E$ is trained to be equivalent to $G^{-1}$.
\begin{equation}
    \label{eq:vgae}
    \min_{G, E} \max_{D} V \left(D, G, E \right) \nonumber
\end{equation} 
These models are optimized alternately with the following objective function with coefficients~$\lambda$s of each $\mathcal{L}$:
\begin{equation}
    \label{eq:loss_all}
    \begin{split}
     V \left(D, G, E \right) =& \, \lambda_{\mathrm{GAN}} \, \mathcal{L}_{\mathrm{GAN}} \left(D, G\right) + 
      \, \lambda_{\mathrm{rec}} \, \mathcal{L}_{\mathrm{rec}} \left(G, E\right) \\
        &+ \lambda_{\mathrm{map}} \, \mathcal{L}_{\mathrm{map}} \left(G\right) 
         + \lambda_{\mathrm{col}} \, \mathcal{L}_{\mathrm{col}} \left(D\right)
    \end{split}
\end{equation} 

\begin{enumerate}
    \item $\mathcal{L}_{\mathrm{GAN}}$: The primary loss function to learn the mapping from the latent space to the joint space.
    \item $\mathcal{L}_{\mathrm{rec}}$: The loss function constrains latent space and joint space so that they can be reconstructed with each other.
    \item $\mathcal{L}_{\mathrm{map}}$: The loss function constrains the transformation from the latent space to the joint space to be smooth.
    \item $\mathcal{L}_{\mathrm{col}}$: The loss function to learn various obstacle situations even if the collision data is a small number on the whole, including non-collision data.
\end{enumerate}

The detail of four $\mathcal{L}$s will be explained in the following Section~\ref{sec:L_GAN} through Section~\ref{sec:L_col}.

\subsubsection{$\mathcal{L}_{\mathrm{GAN}}$: Construction of Latent Expression}
\label{sec:L_GAN}
cGANs are used to construct the mapping from the latent space to the joint space.
In GANs~\cite{goodfellow2014generative}, latent expressions are constructed by training two models, a generator~$G$ and a discriminator~$D$, alternately.
The generator~$G$ creates data variables $\boldsymbol{\theta}_{\mathrm{model}}$ from latent variables $\boldsymbol{z}$.
The discriminator~$D$ estimates whether given variables are a sample from the data set $\boldsymbol{\theta}_{\mathrm{non \mathchar`- col}}$ or a generated sample $\boldsymbol{\theta}_{\mathrm{model}}$ calculated from $\boldsymbol{z}$, which is uniformly sampled from the latent space within $[0,1]$.
That is, for an N-dimensional vector as a latent variable, each of its elements is in the range $[0,1]$.
Since the latent space is a convex space and the boundaries of the latent space can be arbitrarily determined in advance, any point of a line segment connecting any point is in that latent space within a domain of definition.
Furthermore, it is possible to give conditions to the models by introducing a $condition$ variable $\boldsymbol{c}$~\cite{mehdi2014conditional}.
In our case, $\boldsymbol{c}$ is a depth image as the obstacle information.

Fig.~\ref{fig:network} shows the concept of the proposed network model.
Through the generator~$G$, the mapping from the latent space to collision-free joint space is obtained.
The discriminator~$D$ identifies the joint angles, generated joint angles $\boldsymbol{\theta}_{\mathrm{model}}$ by the generator~$G$, and the actual sampled joint angles $\boldsymbol{\theta}_{\mathrm{non \mathchar`- col}}$.
The obstacle information is given as a depth image in condition $\boldsymbol{c}$.
This condition $\boldsymbol{c}$ is connected to the generator~$G$ and the discriminator~$D$ so that when the given obstacle information changes, the correspondence from the latent space to joint space changes.
In other words, our method does not need to prepare a different network for each obstacle, and only one cGANs can support multiple obstacle environments.
The loss function, $\mathcal{L}_{\mathrm{GAN}}$, for training cGANs is shown in  equation~\eqref{eq:loss_gan}.
\begin{equation}
    \label{eq:loss_gan}
    \begin{split}
    \mathcal{L}_{\mathrm{GAN}}(D,& G) = 
        \mathbb{E}_{ 
            \boldsymbol{c} \sim p_{\mathrm{obs}} \left( \boldsymbol{c} \right), \,
            \boldsymbol{\theta} \sim p_{\mathrm{non \mathchar`- col}} \left( \boldsymbol{\theta} | \boldsymbol{c} \right)
        } \left[ 
            \log D \left( \boldsymbol{\theta} , \boldsymbol{c} \right) 
        \right] \\
         & + \mathbb{E}_{
            \boldsymbol{c} \sim p_{\mathrm{obs}} \left( \boldsymbol{c} \right), \,
            \boldsymbol{z} \sim p_{\boldsymbol{z}} \left(\boldsymbol { z }\right)
        } \left[
            \log \left( 
                1 - D \left( G \left( 
                         \boldsymbol{z} , \boldsymbol{c} 
                    \right) , \boldsymbol{c} 
                \right) 
            \right) 
        \right] 
    \end{split}
\end{equation}
\noindent
Where $p_{\mathrm{obs}}(\boldsymbol{c})$ is the distribution of obstacles positions and $p_{\mathrm{non \mathchar`- col}}(\boldsymbol{\theta}|\boldsymbol{c})$ is the distribution of non-collision joint angles which the generator should aim to generate.
$p_{\boldsymbol{z}}(\boldsymbol{z})$ is the uniform distribution in the latent space.

\subsubsection{$\mathcal{L}_{\mathrm{rec}}$: Reconstruction of latent variables and joint angles}
\label{sec:L_rec}
This section describes an objective function that constrains the latent space and joint space so that they can be reconstructed from each other.
The transformation from the latent space to joint space can be trained by $\mathcal{L}_{\mathrm{GAN}}$, but the transformation from the joint space to latent space requires the inverse transformation of the generator~$G^{-1}$.
Therefore, encoder~$E$ is trained as an inverse transformation of generator~$G^{-1}$ and simultaneously learns the reconstruction of the latent space and the joint space so that each point in the joint space and latent space corresponds one-to-one, which means constraining the generator~$G$ to be a single projection.
This loss function, $\mathcal{L}_{\mathrm{rec}}$, is shown in equation~\eqref{eq:loss_rec}.

\begin{equation}
    \label{eq:loss_rec}
    \begin{split}
    \mathcal{L}_{\mathrm{rec}} & (G, E) =
        \mathbb{E}_{ 
            \boldsymbol{c} \sim p_{\mathrm{obs}} \left( \boldsymbol{c} \right), \,
            \boldsymbol{\theta} \sim p_{\mathrm{non \mathchar`- col}} \left( \boldsymbol{\theta} | \boldsymbol{c}  \right)
        } \left[
            \| 
                G \left( E \left(
                    \boldsymbol{\theta}
                        , \boldsymbol{c} 
                     \right)
                    , \boldsymbol{c}
                \right) 
                - \boldsymbol{\theta}
            \|_{2}^{2}
        \right] \\
      & + \mathbb{E}_{ 
        \boldsymbol{c} \sim p_{\mathrm{obs}} \left( \boldsymbol{c} \right), \,
        \boldsymbol{z} \sim p_{\boldsymbol{z}} \left(\boldsymbol { z }\right)
    } \left[
        \| 
            E \left( G \left(
                \boldsymbol{z}
                    , \boldsymbol{c} 
                 \right)
                , \boldsymbol{c}
            \right) 
            - \boldsymbol{z}
        \|_{2}^{2}
    \right] \\
    \end{split}
\end{equation}
\subsubsection{$\mathcal{L}_{\mathrm{map}}$: Specifying the Map from the Latent Space to Joint Space}
\label{sec:L_map}
We will describe in this section how to map from the latent space to joint space such that arbitrary planned paths in the latent space are smooth in joint space for robot arms.
For planning using a robot arm, the mapping from the latent space to joint space has to be continuous without ``twists'', ``distortions'', and rapid changes.
In order to achieve this, the following two things are performed:
a) The number of dimensions for latent variables is matched to the number of robot joints; each latent variable is mapped to represent each joint, and the normalized ranges of latent variables and joint angles are aligned.
b) The generator~$G$ is trained to output $\boldsymbol{\theta}$ when the latent variables $\boldsymbol{z}=\boldsymbol{\theta}$ are given as input of the generator~$G$.
The procedure for calculating $\mathcal{L}_{\mathrm{map}}$ is as follows three steps.
1) The obstacle condition is determined by sampling from the distribution of obstacle positions: $\boldsymbol{c} \sim p_{\mathrm{obs}}\left( \boldsymbol{c} \right)$. 
2) Non-collision joint angles is sampled under this condition $\boldsymbol{c}$: $\boldsymbol{\theta} \sim p_{\mathrm{non \mathchar`- col}} \left( \boldsymbol{\theta} | \boldsymbol{c} \right)$.
3) The output of the generator is constrained to be $\boldsymbol{\theta}$ itself when this $\boldsymbol{\theta}$ is input as $\boldsymbol{z}=\boldsymbol{\theta}$.
The generator is trained to be close to an identity map for non-collision joint angles under condition $\boldsymbol{c}$.
Colliding joint angles are not sampled in $\mathcal{L}_{\mathrm{map}}$.
The constraint is not added to the joint that collides with the obstacles so that the constructed map is allowed to be distorted to avoid collisions.
The loss function, $\mathcal{L}_{\mathrm{map}}$, for training cGANs is shown in equation~\eqref{eq:loss_im}.
\begin{equation}
    \label{eq:loss_im}
    \mathcal{L}_{\mathrm{map}}(G) = 
        \mathbb{E}_{ 
            \boldsymbol{c} \sim p_{\mathrm{obs}} \left( \boldsymbol{c} \right), \,
            \boldsymbol{\theta} \sim p_{\mathrm{non \mathchar`- col}} \left( \boldsymbol{\theta} | \boldsymbol{c}  \right)
        } \left[
            \| 
                G \left( \boldsymbol{z} = \boldsymbol{\theta} , \boldsymbol{c} 
                \right) - \boldsymbol{\theta}
            \|_{2}^{2}
        \right] 
\end{equation}
\subsubsection{$\mathcal{L}_{\mathrm{col}}$: Adaptability to Multiple Obstacle Conditions}
\label{sec:L_col}
In this section, we describe how to adapt to various obstacle conditions.
Even though collision-free mapping from the latent space to joint space is trained by equation~\eqref{eq:loss_gan}, the network has a risk of mistaking collision points for non-collision points, and vice versa since the number of non-collision data points is much smaller than those with collisions.
The collision joints must be explicitly included in the equation during training.
The loss function, $\mathcal{L}_{\mathrm{col}}$, shown in equation~\eqref{eq:loss_col} is introduced in order to provide the data of the collision joints to the discriminator~$D$.
\begin{equation}
    \label{eq:loss_col}
    \mathcal{L}_{\mathrm{col}}(D) = 
        \mathbb{E}_{ 
            \boldsymbol{c} \sim p_{\mathrm{obs}} \left( \boldsymbol{c} \right), \,
            \boldsymbol{\theta} \sim p_{\mathrm{col}} \left( \boldsymbol{\theta} | \boldsymbol{c} \right)
        } \left[
            \log \left( 
                1 - D \left( \boldsymbol{\theta} , \boldsymbol{c} 
                      \right) 
                 \right) 
        \right] 
\end{equation}
\noindent
Where $p_{\mathrm{col}}(\boldsymbol{\theta} | \boldsymbol{c} )$ is the distribution of colliding joint angles, including self-collision and collision with obstacles, which the generator~$G$ should thus refrain from generating.
The discriminator~$D$ is trained to output $0$ for collision joints and $1$ for collision-free joints for each obstacle.
Furthermore, the generator~$G$ is trained to acquire a distribution to make the discriminator~$D$ output $1$, as we are trying to obtain a distribution for collision-free space.
\subsection{Planning}
\label{sec:path_planning}
In this section, we will describe the planning method.
Section~\ref{sec:path_selection} describes how to generate various optimal trajectories for different purposes, and Section~\ref{sec:path_correction} explains how to guarantee collision avoidance with obstacles since learning methods alone cannot wholly avoid them.
Then, Section~\ref{sec:computational_complexity} explains computational complexity.
\subsubsection{Optimal Path Trajectory Generation}
\label{sec:path_selection}
In our method, the mapping and planning phases are separated, unlike traditional path planners in joint space.
Moreover, any path planner can be used in the trained latent space (where any point is collision-free) without considering obstacles since there are none in the latent space. 
Therefore, it makes our method highly \textbf{\emph{customizable}}.
As optimization methods for any optimization criterion, we can not only use any discrete optimization methods, such as $A^{*}$~\cite{hart1968formal} on any graphs in the latent space but also continuous optimization methods, such as Adam~\cite{adam}, thanks to the differentiable nature of the generator~$G$.
The computational cost is also lower since collision check calculations are no longer necessary, making our method also \textbf{\emph{scalable}}.

As shown in Fig.~\ref{fig:method_transform}, when the start joint angles $\boldsymbol{\theta}_s$ and the goal joint angles $\boldsymbol{\theta}_g$ are given, the corresponding latent variables are found by $ \boldsymbol{z}_s = E(\boldsymbol{\theta}_s, \boldsymbol{c})$, $ \boldsymbol{z}_g = E(\boldsymbol{\theta}_g, \boldsymbol{c})$. 
Considering $\boldsymbol{z}_{s:g}$ as a path connecting these in the latent space, the collision-free path is obtained as $\boldsymbol{\theta}_{s:g} = G(\boldsymbol{z}_{s:g}, \boldsymbol{c})$.
$\boldsymbol{z}_{s:g}$ can be determined arbitrarily within the latent space, and in the simplest case, it can be connected by a straight line.
Taking advantage of the differentiability of the generator~$G$, the path in the latent space can be calculated by optimizing the cost function $\mathcal{L}_{\mathrm{opt}}$ to satisfy the objective using the following equation:

\begin{equation}
    \label{eq:cost}
     \begin{split}
    \mathcal{L}_{\mathrm{opt}} & \, = f(G,\boldsymbol{z}_{s:g}, \boldsymbol{c}) \\
    \hat{\boldsymbol{z}}_{s:g} & \, = \argmin_{\boldsymbol{z}_{s:g}} \mathcal{L}_{\mathrm{opt}}
     \end{split}
\end{equation}
\noindent

There are a variety of cost functions depending on the objectives.
For example, $\mathcal{L}_{\mathrm{opt}}$ can be minimization of the sum of squares of velocities $\mathcal{L}_{\mathrm{opt}} = \objFunc{v}$, 
minimization of acceleration $ \mathcal{L}_{\mathrm{opt}} = \objFunc{a}$, 
and minimization of jerk $\mathcal{L}_{\mathrm{opt}} = \objFunc{j}$, where $\boldsymbol{v}_{t} = \boldsymbol{\theta}_{t} - \boldsymbol{\theta}_{t-1}$, $\boldsymbol{a}_{t} = \boldsymbol{v}_{t} - \boldsymbol{v}_{t-1}$, and $\boldsymbol{j}_{t} = \boldsymbol{a}_{t} - \boldsymbol{a}_{t-1}$.
Also, the path can be optimized by combining them as the following equation:
\begin{equation}
    \label{eq:opt}
    \begin{split}
    \mathcal{L}_{\mathrm{opt}} = \sum_{t} \|\boldsymbol{v}_{t}\| ^{2}_{2} + \alpha\sum_{t} \|\boldsymbol{a}_{t}\| ^{2}_{2} + \beta\sum_{t} \|\boldsymbol{j}_{t}\| ^{2}_{2}
     \end{split}
\end{equation}
\subsubsection{Collision Avoidance Guarantee (CAG)}
\label{sec:path_correction}
The learning method described above does not guarantee 100\% obstacle avoidance.
Therefore, the trajectory obtained by the generator~$G$ is checked to ensure that it does not collide with any obstacles.
Moreover, the trajectory is modified to guarantee collision avoidance if a colliding posture is found.
If the path includes a colliding posture, the non-colliding posture before and after the collision trajectory is modified using existing planning methods.
In this study, we use RRT Connect~\cite{kuffner2000rrt}.
\subsubsection{Computational Complexity}
\label{sec:computational_complexity}
Our proposed method (Ours w/o collision avoidance guarantee (CAG)) performs forward calculations for the path's number of iterations $I_1$, and the computational complexity is $O(I_1)$.
Assume that the proposed method's number of iterations $I_1$ is sufficiently large so that the robot's degrees of freedom can be ignored.
We do not consider path optimization here.
Our proposed method uses collision-free latent space and the computational complexity does not depend on the complexity of obstacles.
In the simplest case, the path is a straight line connecting from the start to the goal in the latent space, and $I_1$ is obtained by dividing the straight line distance $L_1$ in latent space by the distance $D_1$ per step, $O(I_1=L_1/D_1)$.

The computational complexity of the RRT is $O({I_2}^2)$ for the number of iterations $I_2$~\cite{rodriguez2013blind, svenstrup2011minimising}. 
Assume that the number of iterations $I_2$ is sufficiently large so that the robot's degrees of freedom can be ignored.
Since RRT's calculation is performed in space with obstacles, unlike the proposed method, the more obstacles there are, the more complex the trajectory from the start to the goal will be, and the more trajectories will collide with obstacles during the RRT calculation, the more iterations $I_2$ will be required.
\begin{figure}[t]
    \centering
    \includegraphics[width=0.95\columnwidth]{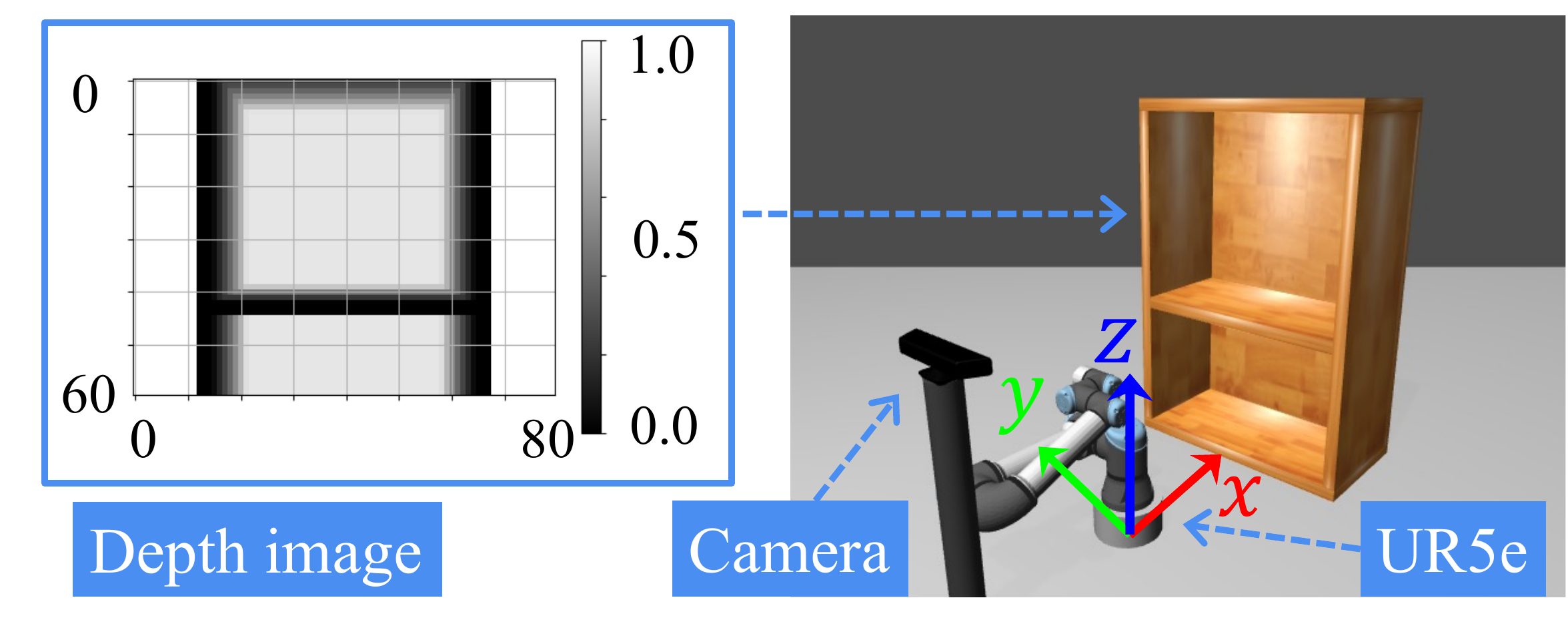}
    \caption{Experiment environment and depth image of obstacle.}
    \label{fig:experiment_robot}
\end{figure}

\begin{figure}[t]
    \centering
    \includegraphics[width=0.95\columnwidth]{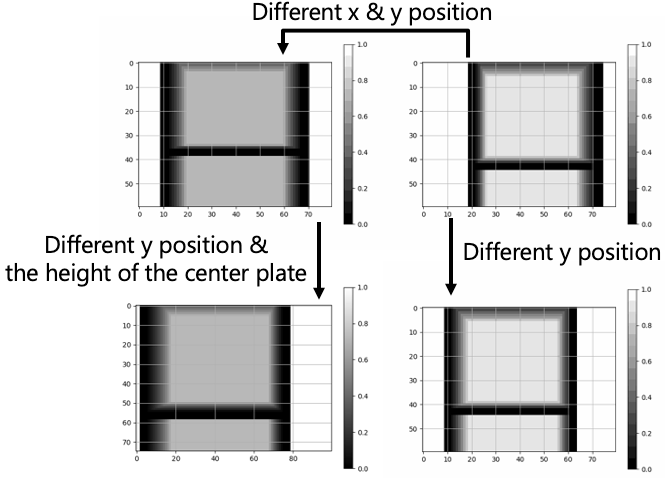}
    \caption{Examples of a depth image with different $x$ and $y$ positions and the height of the center plate. The shelf's color in-depth image changes as the shelf's position changes in the $x$-coordinate. The test evaluation was performed with untrained $x$, $y$, and the height of the center plate.}
    \label{fig:examples_of_depth_image}
\end{figure}

\section{Experimental Setup}
\label{sec:experiment}
We performed experiments using a Gazebo simulation and an actual 6-DoF UR5e robot arm.
In learning-based methods using robotic arms in 3-D space, objects like pillars or shelves are often used as obstacles.
We use a shelve for practical tasks.
We set up an environment with a shelf in front of the robot as an obstacle and evaluated the system's performance under multiple conditions by changing the obstacle position of the shelf and the height of the center plate of the shelf (Fig.~\ref{fig:experiment_robot} and Fig.~\ref{fig:examples_of_depth_image}).
\subsection{Data Collection}
\subsubsection{Obstacle Information}
\label{sec:obstacle_information}
The size of the placed shelf as an obstacle is $120~\mathrm{cm}$ in height, $80~\mathrm{cm}$ in width, and $40~\mathrm{cm}$ in depth.
The origin of the coordinate system is the floor just below the root of the robot, with the forward direction in the positive $x$-axis, the left side in the positive $y$-axis, and the upper side in the positive $z$-axis.
The robot is placed on a $10~\mathrm{cm}$ high pedestal.
The shelf is placed in various locations, which are divided into 5 positions by $x\in [60~\mathrm{cm}, 70~\mathrm{cm}]$ and 9 positions by $y\in [-10~\mathrm{cm}, 10~\mathrm{cm}]$, and the shelf is located so that the training data and test data alternated (Fig.~\ref{fig:examples_of_depth_image}).
Similarly, the height of the board is divided into 11 parts in the range $z \in [50~\mathrm{cm}, 60~\mathrm{cm}]$ so that the training data and test data alternate.
Therefore, there are $23\,(\mathrm{positions}) \times 6\,(\mathrm{heights}) = 138$ different types of conditions for training data, and $22\,(\mathrm{positions}) \times 5\,(\mathrm{heights}) = 110$ different types of condition for test data. 

We used a depth image taken from behind the robot for the condition $\boldsymbol{c}$ that indicates the obstacle information.
The depth image is taken in such a way that the robot is not included in the image, and one depth image corresponds to each obstacle condition.
The depth image is a one-channel image with a size of $60 \times 80$ pixels.
The shelf is placed in the $ [40~\mathrm{cm}, 90~\mathrm{cm}]$ range on the $x$-axis, and the depth information is normalized by $[0, 1]$ in that range.
\subsubsection{Robot Information}
The datasets of cGANs consists of 6 joint angles $\boldsymbol{\theta}=(\theta_1, \theta_2, \theta_3, \theta_4, \theta_5, \theta_6)$ that represents the robot's posture.
The ranges of the joint angles are $\theta_1 \in [-90^\circ, 90^\circ]$, $\theta_2 \in [-120^\circ, 120^\circ]$, $\theta_{3\mathchar`-6} \in [-180^\circ, 180^\circ]$.
The latent variables $\boldsymbol{z}$ are 6-dimensional because the robot has 6-DoF, and a uniform distribution in the range $[0, 1]$ is used.

We collected a total of 50,000 collision and non-collision data points of robot postures for various obstacle conditions by random sampling and used them for training.
The percentage of collision data is about 43\% of the total, which includes self-collision and floor collision (about 37\%) and shelf collision (about 9\%).
In some cases, both collisions co-occur, so the total exceeds 43\%.
Min-Max scaling was applied to each $\theta_i$, normalized to the $[0, 1]$ range.
\begin{table}[t]
    \centering
    \begin{threeparttable}[t]
    \caption{Network design}
    \label{table:network_design}
        \begingroup
        \scalefont{0.85}
        \begin{tabular}{c|c||c c c c c}
            \hline 
            \multicolumn{1}{c|}{} &  Layer & In & Out & \begin{tabular}{c}Filter\\size\end{tabular} & \begin{tabular}{c}Norma-\\-lization\end{tabular} & \begin{tabular}{c}Activation\\function\end{tabular}\\
            \hline
            \multirow{12}{*}{Conv\tnote{1}} 
                & \nth{1} conv. & 1  & 4  & (3,3) & BN & Leaky ReLU \\
                & \nth{2} conv. & 4  & 4  & (3,3) & BN & Leaky ReLU \\
                & AvgPool1 & 4  & 4  & (2,2) & -  & - \\
                & \nth{3} conv. & 4  & 8  & (3,3) & BN & Leaky ReLU \\
                & \nth{4} conv. & 8  & 8  & (3,3) & BN & Leaky ReLU \\
                & AvgPool2 & 8  & 8  & (2,2) & -  & - \\
                & \nth{5} conv. & 8  & 16 & (3,3) & BN & Leaky ReLU \\
                & AvgPool3 & 16 & 16 & (2,2) & -  & - \\
                & $\mathrm{FC_{obs \mathchar`- 0}}$           & 1120 & 1024 \tnote{2} & - & BN & Leaky ReLU \\
                & $\mathrm{FC_{obs \mathchar`- 1}}$ & 1024 \tnote{2} & 1024 \tnote{3} & - & BN & Leaky ReLU \\
                & $\mathrm{FC_{obs \mathchar`- 2}}$ & 1024 \tnote{2} & 1024 \tnote{3} & - & BN & Leaky ReLU \\
                & $\mathrm{FC_{obs \mathchar`- 3}}$ & 1024 \tnote{2} & 1024 \tnote{3} & - & BN & Leaky ReLU \\
            \hline
            \multirow{8}{*}{$D$} 
                & \nth{1} FC & 6    & 256            & - & -  & Leaky ReLU \\
                & \nth{2} FC & 256  & 512            & - & SN & Leaky ReLU \\
                & \nth{3} FC & 512  & 1024 \tnote{3} & - & SN & Leaky ReLU \\
                & \nth{4} FC & 1024 & 1024 \tnote{3} & - & SN & Leaky ReLU \\
                & \nth{5} FC & 1024 & 1024 \tnote{3} & - & SN & Leaky ReLU \\
                & \nth{6} FC & 1024 & 1024           & - & SN & Leaky ReLU \\
                & \nth{7} FC & 1024 & 1024           & - & SN & Leaky ReLU \\
                & \nth{8} FC & 1024 & 1              & - & -  & Linear \\
            \hline
                & \nth{1} FC & 6    & 256            & - & -  & Leaky ReLU \\
                & \nth{2} FC & 256  & 512            & - & SN & Leaky ReLU \\
                & \nth{3} FC & 512  & 1024 \tnote{3} & - & SN & Leaky ReLU \\
            $G$ & \nth{4} FC & 1024 & 1024 \tnote{3} & - & SN & Leaky ReLU \\
            \&\tnote{4}& \nth{5} FC & 1024 & 1024 \tnote{3} & - & SN & Leaky ReLU \\
            $E$ & \nth{6} FC & 1024 & 1024           & - & SN & Leaky ReLU \\
                & \nth{7} FC & 1024 & 1024           & - & SN & Leaky ReLU \\
                & \nth{8} FC & 1024 & 6              & - & -  & Linear \\
            \hline
        \end{tabular}
        \endgroup
        \begin{tablenotes}
            \scriptsize{
            \item[1] $G$, $D$, and $E$ have independent feature extraction units.
            \item[2] The output of $\mathrm{FC_{obs \mathchar`- 0}}$ is the input to $\mathrm{FC_{obs \mathchar`- 1}}$, $\mathrm{FC_{obs \mathchar`- 2}}$, and $\mathrm{FC_{obs \mathchar`- 3}}$, respectively.
            \item[3] The element-wise product of the output of $\mathrm{FC_{obs \mathchar`- 1}}$ and the output of \nth{3} FC is the input to the next layer. 
            The same process applies to the outputs of $\mathrm{FC_{obs \mathchar`- 2}}$ and \nth{4} FC, and to the outputs of $\mathrm{FC_{obs \mathchar`- 3}}$ and \nth{5} FC.
            \item[4] $G$ and $E$ have the same structure but independent parameters.
            }
    \end{tablenotes}
    \end{threeparttable}
\end{table}

\subsection{Network Design}
Our network model is composed of $G$, $D$, and $E$ with fully connected layers, and each network includes a two-dimensional convolutional layer (conv.) as a feature extraction unit for conditions~$\boldsymbol{c}$ (Fig.~\ref{fig:network}).
The network design details are shown in Table~\ref{table:network_design}.
For learning stabilization, batch normalization~\cite{batch_norm} are spectral normalization~\cite{sngan} were used.
Our network model is implemented with PyTorch.
Training is conducted on a machine equipped with Intel Core i7-11700F@2.50GHz CPU and GeForce RTX 3070, resulting in about 3 to 4 days of training time.

We describe $\lambda$s, which are the coefficients of each loss function in the equation~\eqref{eq:loss_all}.
They are set as $\lambda_{\mathrm{GAN}}=1$, $\lambda_{\mathrm{rec}}=100$, 
$\lambda_{\mathrm{map}}=10$, and $\lambda_{\mathrm{col}}=100$.
When the distance between the robot and the obstacle is less than $ 10~\mathrm{cm}$, $\lambda_{\mathrm{rec}}=0$ and $\lambda_{\mathrm{map}}=0$ are used.
If the robot collides with the shelf, $\lambda_{\mathrm{col}}=1000$.
This aims to increase safety by ensuring the distance to collision, and generator, $G$, was trained to exclude postures close to obstacles.
\subsection{Settings for the Comparison Methods}
The Python implementation in ROS was used for the RRT Connect~\cite{kuffner2000rrt} used in the proposed method and for the RRT~\cite{lavalle1998rapidly} and RRT Connect used for comparison with the proposed method.
We used the default parameters of the motion planning framework MoveIt!.

\section{Experiment Results}
\label{sec:result}
We will confirm the \rnum{2}~)~\textbf{\emph{Adaptability}} to various obstacles by evaluating the constructed mapping in section~\ref{sec:evaluate_mapping}. 
Next, \rnum{1}~)~\textbf{\emph{Customizability}} will be confirmed by planning on several optimization criteria in section~\ref{sec:result_path_select}, and \rnum{3}~)~\textbf{\emph{Scalability}} will be confirmed by comparison with other planning methods in section~\ref{sec:result_path_time} and collision avoidance guarantee will be verified.
Finally, we show results with an actual robot, UR5e, in section~\ref{sec:result_ur5e}.
Also, pre-experiments were conducted in a 2-D environment with various shapes and numbers of obstacles. See Appendix for details.
\begin{figure}[t]
    \centering
    \includegraphics[width=0.95\columnwidth]{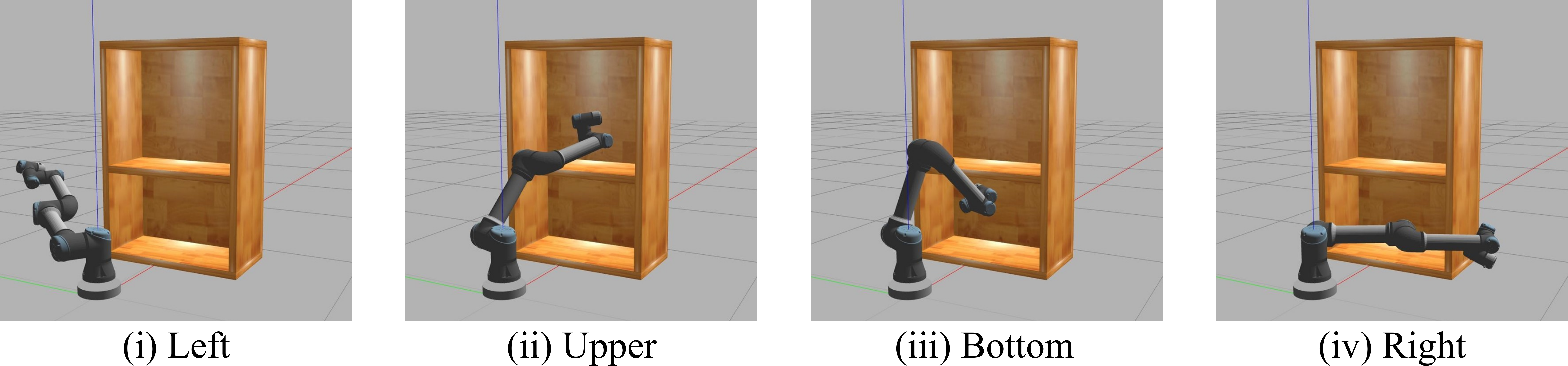}
    \caption{Four regions of start and goal for trajectory generation: \emph{left}, \emph{upper}, \emph{bottom}, and \emph{right} relative to the shelf.
    Random postures were set as the start and the goal from two regions.}
    \label{fig:shelf_robot}
\end{figure}

\begin{table}[t]
    \begin{minipage}{1.0\columnwidth}
        \tblcaption{The success rate of planning. Planning was performed by connecting straight lines in latent space for a given start and goal.
        The results of brackets include those that could generate trajectories without collisions with obstacles but failed to reconstruct.
        }
        \label{table:result_col_success} 
        \centering
        \begingroup
            \scalefont{1.0}
            \begin{center}
           \scalebox{1.00}{
                \begin{tabular}{c c c c c }
                \hline 
                 & $\mathcal{L}_{\mathrm{map}}$ 
                 & $\mathcal{L}_{\mathrm{col}}$ 
                 & \multicolumn{1}{c}{Dataset} 
                 & \begin{tabular}{c}Path Success\\ Rate [\%]\end{tabular} \\
                 
                \hline
                \multirow{2}{*}{Ours} &
                \multirow{2}{*}{w} &
                \multirow{2}{*}{w} &
                Train & 72.7 (85.3) \\
                & & & 
                Test  & 70.9 (89.4) \\
                
                \hline
                \multirow{2}{*}{w/o $\mathcal{L}_{\mathrm{map}}$} & 
                \multirow{2}{*}{w/o} & 
                \multirow{2}{*}{w} & 
                Train & 22.9 (24.2) \\
                & & & 
                Test  & 15.8 (20.0) \\
                \hline
                \multirow{2}{*}{w/o $\mathcal{L}_{\mathrm{col}}$} & 
                \multirow{2}{*}{w} & 
                \multirow{2}{*}{w/o} & 
                Train & 17.6 (31.9)  \\
                 & & & 
                Test  & 17.0 (32.1) \\
                \hline 
                \end{tabular}
                }
            \end{center}
    \endgroup
    \end{minipage}
\end{table}

\subsection{Evaluation of Adaptability to Various Obstacles}
\label{sec:evaluate_mapping}
We evaluated the accuracy of the mapping acquired as $G$ and $E$ by evaluating the success rate of planning to verify \rnum{2}~)~\textbf{\emph{Adaptability}}.
For the evaluation, random postures were set as the start and the goal from two different regions among the regions whose end-effector positions were \emph{left}, \emph{upper}, \emph{bottom}, and \emph{right} relative to the shelf (Fig.~\ref{fig:shelf_robot}).
Then, for various obstacle conditions as described in Section~\ref{sec:obstacle_information}, we let the robot arm's end-effector plan a path to cross the shelf board using the above start and goal posture.
Note that the posture of the robot arm at the start and goal is at least 5~$\mathrm{cm}$ away from the collision.
For each obstacle condition, three paths are generated; $138\,conditions \times 3\,paths = 414\,paths$ are evaluated for train datasets, and $110\,conditions \times 3\,paths = 330\,paths$ are evaluated for test datasets.
Note that the obstacle conditions used in the test are not used in training.
Although there are several possible paths in the latent space, we use the path that connects the two points by a straight line in the latent space after calculating $ \boldsymbol{z}_s = E(\boldsymbol{\theta}_s, \boldsymbol{c})$ and $ \boldsymbol{z}_g = E(\boldsymbol{\theta}_g, \boldsymbol{c})$ using the encoder $E$ for the start and goal postures $\boldsymbol{\theta}_s$ and $\boldsymbol{\theta}_g$, respectively.
The length of $\boldsymbol{z}_{s:g}$ is fixed at 200 steps.

Success in planning means that the generated path does not contain any collision postures and that the start and goal positions are reached.
The method for determining whether the start and goal positions have been reached is to calculate the Euclidean distance difference of $(x,y,z)$ between the end-effector position calculated from the given start and goal postures and the reconstructed start and goal postures from the latent variables generated by planning as follows:
\begin{equation}
    \label{eq:distance}
     \begin{split}
    \| FK(\boldsymbol{\theta}_{\mathrm{rec}})  - FK(\boldsymbol{\theta}_{ \mathrm{target}}) \| < \epsilon
    \end{split}
\end{equation}
where $FK()$ is forward kinematics to calculate end-effector position  from $\boldsymbol{\theta}$,  $\boldsymbol{\theta}_{\mathrm{rec}}=
 G \left(
     E(\boldsymbol{\theta}_{\mathrm{target}}, \boldsymbol{c}
 ), \boldsymbol{c} \right)$, and $\epsilon=5.0~\mathrm{cm}$.

In this experiment, we verify the effectiveness of each loss function of the proposed method, which consists of four loss functions $\mathcal{L}$s, as shown in equation~\eqref{eq:loss_all}.
Since $\mathcal{L}_{\mathrm{GAN}}$ and $\mathcal{L}_{\mathrm{rec}}$ are the minimum required for the training of the model, the following three conditions are used to examine the effectiveness of the other $\mathcal{L}$s: (a) Our proposed method, (b) without $\mathcal{L}_{\mathrm{map}}$ from equation~\eqref{eq:loss_all}, and (c) without $\mathcal{L}_{\mathrm{col}}$ from equation~\eqref{eq:loss_all}.
Note that the collision avoidance guarantee (CAG) is not used in the planning of any method.

Table~\ref{table:result_col_success} shows the results of the experiment.
The success rates shown in brackets in Table~\ref{table:result_col_success} include those that could generate trajectories without collisions with obstacles, but did not satisfy equation~\eqref{eq:distance}, i.e., failed to reconstruct.
The success rate drops significantly without either $\mathcal{L}_{\mathrm{map}}$ or $\mathcal{L}_{\mathrm{col}}$.
Since the path success rate is low even when failure of reconstruction is considered, it can be said that a significant rate of collisions with obstacles occurs in trajectory.

We also confirmed that the success rate decreased by 18.5\% when a depth image different from actual obstacle information was provided to the condition.
From the results, we confirmed that the generated paths varied depending on the conditions.
The proposed method has a high success rate of more than 70\% for both the training and test dataset, indicating that it can generate trajectories even under untrained obstacle conditions.
In other words, the \textbf{\emph{adaptability}} of the proposed method to various obstacle conditions is verified.
\begin{table}[tb]
    \centering
    \caption{Evaluation of the optimized trajectory}
    \label{table:optimal-path}
    \scalebox{1.0}{
        \begin{tabular}{cccc}
            \hline
            \multicolumn{1}{c}{Target to optimize} & $\objFunc{v}$      & $\objFunc{a}$    & $\objFunc{j}$ \\
            
            \hline \hline
            w/o Opt. & 
            1.16 $\pm$ 0.21 & 0.20 $\pm$ 0.12 & 0.38 $\pm$ 0.27 \\
            \hline
            
            $\boldsymbol{v}$ &
            $\boldsymbol{0.75 \pm 0.25}$ & 0.055 $\pm$ 0.012 & 0.11 $\pm$ 0.032 \\
            
            $\boldsymbol{a}$ &
            1.06 $\pm$ 0.29 & $\boldsymbol{0.031 \pm 0.014}$ & 0.033 $\pm$ 0.024 \\
            
            $\boldsymbol{j}$ & 
            1.17 $\pm$ 0.23 & 0.052 $\pm$ 0.017 & $\boldsymbol{0.026 \pm 0.010}$ \\
            
            Mix of $\boldsymbol{v}, \boldsymbol{a}, \boldsymbol{j}$ &
            0.92 $\pm$ 0.25 & 0.042 $\pm$ 0.011 & 0.049 $\pm$ 0.016 \\
            \hline
        \end{tabular}
    }
\end{table}

\begin{figure}[t]
    \centering
    \includegraphics[width=0.95\columnwidth]{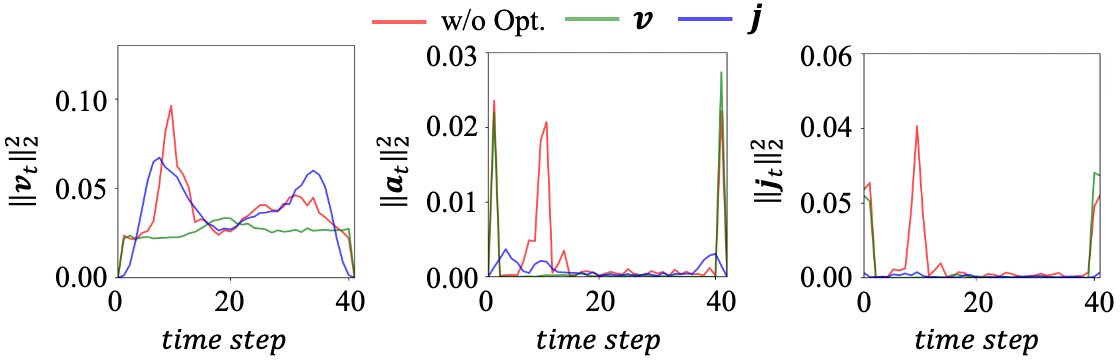}
    \caption{The values $\boldsymbol{v}, \boldsymbol{a}, \boldsymbol{j}$ of before and after optimization.
    }
    \label{fig:optimal-path}
\end{figure}

\subsection{Planning on Arbitrary Optimization Criteria}
\label{sec:result_path_select}
In this section, we verify \rnum{1}~)~\textbf{\emph{Customizability}}.
The proposed method can generate multiple (in-finite) paths.
The method for determining paths in the latent space is not limited to just connecting the start and the goal linearly but can be any path/trajectory planner.
As examples of optimization, Table~\ref{table:optimal-path} shows the values of the trajectories before and after optimization when velocity~$\boldsymbol{v}$, acceleration~$\boldsymbol{a}$, and jerk~$\boldsymbol{j}$ are minimized, and combinations are performed as described in Section~\ref{sec:path_selection}.
The hyper-parameters in the combination of optimization were set to $\alpha=0.5$ and $\beta=0.5$ in equation~\eqref{eq:opt}.
The 330 trajectories (110 untrained conditions $\times$ three pairs of start and goal) from the test dataset used in Section~\ref{sec:evaluate_mapping} were used as the trajectories before optimization.
Only those of these trajectories that avoided obstacles before and after optimization were used to generate Table~\ref{table:optimal-path} results.
The results of guaranteeing obstacle avoidance will be shown in Section~\ref{sec:result_path_time}.
From Table~\ref{table:optimal-path}, it can be confirmed that the value subjected to each optimization is the smallest.
These optimizations took approximately $2~\mathrm{s}$ to $10~\mathrm{s}$, and latent variables were updated from 500 to a maximum of 2,500 iterations.

As an example of trajectory optimization, Fig.~\ref{fig:optimal-path} shows the values of $\boldsymbol{v}$, $\boldsymbol{a}$, and $\boldsymbol{j}$ before optimization, which is just a straight line in the latent space, and the trajectory after optimization by velocity minimization and jerk minimization, respectively.
The trajectory before optimization has some parts where the velocity and jerk change suddenly.
By optimizing the trajectory with velocity minimization, the velocity of the entire trajectory is suppressed.
In the trajectory optimized by jerk minimization, the sudden stops and starts at the start and goal are moderated.
Since the values of the graphs generated by each optimization are different, we can say that different trajectories were generated by the optimization.

These results show that planning can be performed using arbitrary optimization criteria, which demonstrates the \textbf{\emph{customizability}} of our proposed method.
\begin{table}[t]
    \centering
    \begin{threeparttable}[t]
    \caption{Success rate and execution time for planning\tnote{1}}
    \label{table:result_path} 
        \begingroup
        \scalefont{0.85}
            \begin{tabular}{c c c c c}
                \hline 
                \multirow{2}{*}{Start-Goal} & \multirow{2}{*}{Method} & Success & Planning  & Path \\
                 & &                           Rate [\%] & Time [$\mathrm{ms}$] & Length [$\mathrm{m}$] \\
                \hline
                \multirow{5}{*}{Left-Upper} &
                Ours w/o CAG & 68.3 & 5.95 $\pm$ 0.16 & 1.48 $\pm$ 0.21 \\
                & Ours (only CAG) & 100.0 & 114.35 $\pm$ 90.10 & 1.73 $\pm$ 0.50 \\
                & Ours & 100.0 & 40.32 $\pm$ 71.02 & 1.56 $\pm$ 0.34 \\
                & RRT & 100.0 & 419.16 $\pm$ 1305.27 & 2.03 $\pm$ 0.59 \\
                & RRT Connect & 100.0 & 119.16 $\pm$ 2.54 & 2.26 $\pm$ 0.88  \\
                \hline
                \multirow{5}{*}{Left-Bottom} &
                Ours w/o CAG & 70.2 & 5.95 $\pm$ 0.17 & 1.91 $\pm$ 0.36 \\
                & Ours (only CAG) & 100.0 & 6.02 $\pm$ 0.25 & 1.97 $\pm$ 0.50 \\
                & Ours & 100.0 & 5.97 $\pm$ 0.20 & 1.93 $\pm$ 0.40 \\
                & RRT & 100.0 & 473.35 $\pm$ 1144.82 & 2.77 $\pm$ 1.47 \\
                & RRT Connect & 100.0 & 126.48 $\pm$ 7.71 & 2.43 $\pm$ 0.97  \\
                \hline
                \multirow{5}{*}{Left-Right} &
                Ours w/o CAG & 81.8 & 5.94 $\pm$ 0.15 & 3.00 $\pm$ 0.39 \\
                & Ours (only CAG) & 100.0 & 35.53 $\pm$ 54.79 & 2.78 $\pm$ 0.43 \\
                & Ours & 100.0 & 11.32 $\pm$ 24.94 & 2.96 $\pm$ 0.40 \\
                & RRT & 100.0 & 201.96 $\pm$ 254.73 & 2.81 $\pm$ 0.80 \\
                & RRT Connect & 100.0 & 125.20 $\pm$ 2.41 & 3.14 $\pm$ 1.46  \\
                \hline
                \multirow{5}{*}{Upper-Bottom} &
                Ours w/o CAG & 69.4 & 5.95 $\pm$ 0.16 & 1.11 $\pm$ 0.27 \\
                & Ours (only CAG) & 100.0 & 89.17 $\pm$ 149.75 & 1.42 $\pm$ 0.78 \\
                & Ours & 100.0 & 31.45 $\pm$ 90.07 & 1.20 $\pm$ 0.50 \\
                & RRT & 98.4 & 1264.04 $\pm$ 3315.85 & 1.88 $\pm$ 1.05 \\
                & RRT Connect & 100.0 & 128.10 $\pm$ 30.49 & 2.03 $\pm$ 0.90  \\
                \hline
                \multirow{5}{*}{Upper-Right} &
                Ours w/o CAG & 73.2 & 5.94 $\pm$ 0.13 & 1.64 $\pm$ 0.32 \\
                & Ours (only CAG) & 100.0 & 76.23 $\pm$ 59.47 & 1.37 $\pm$ 0.56 \\
                & Ours & 100.0 & 24.77 $\pm$ 43.44 & 1.57 $\pm$ 0.41 \\
                & RRT & 100.0 & 377.65 $\pm$ 651.56 & 1.78 $\pm$ 0.69 \\
                & RRT Connect & 100.0 & 118.71 $\pm$ 2.22 & 1.93 $\pm$ 0.71  \\
                \hline
                \multirow{5}{*}{Bottom-Right} &
                Ours w/o CAG & 66.3 & 5.92 $\pm$ 0.12 & 1.91 $\pm$ 0.48 \\
                & Ours (only CAG) & 100.0 & 112.90 $\pm$ 274.59 & 2.65 $\pm$ 1.19 \\
                & Ours & 100.0 & 42.02 $\pm$ 165.55 & 2.16 $\pm$ 0.86 \\
                & RRT & 97.5 & 854.78 $\pm$ 2119.77 & 2.44 $\pm$ 1.51 \\
                & RRT Connect & 100.0 & 136.42 $\pm$ 78.67 & 2.16 $\pm$ 1.17  \\
                \hline
            \end{tabular}
            \endgroup
            \begin{tablenotes}
                \scriptsize{
                    \item[1] The execution time is calculated only for planning and does not include the GPU transfer time (about $10~\mathrm{ms}$)
                }
            \end{tablenotes}
    \end{threeparttable}
\end{table}

\subsection{Comparison of the Planning Times and Collision Avoidance Guarantee}
\label{sec:result_path_time}
Here, we investigate \rnum{3}~)~\textbf{\emph{Scalability}} and collision avoidance guarantee (CAG).
We evaluated the success rate and computation time of the proposed method (which used CAG by RRT Connect if a collision occurred when planning with cGANs).
The 330 trajectories (110 untrained conditions $\times$ three pairs of start and goal) described in Section~\ref{sec:evaluate_mapping} are used. 
These trajectories include different start and goal pairs for each of the untrained obstacle conditions with different shelf positions and different heights of the board (Fig.~\ref{fig:examples_of_depth_image}).
In addition to our method without CAG, the model-based planning methods RRT and RRT Connect were used for comparison.
Since the trajectories generated by RRT and RRT Connect are different each time, three trials were conducted with the same start, goal, and obstacle conditions.
As well as the proposed method used RRT Connect; therefore, three trials were conducted.
To investigate the computation time when RRT Connect is used for the CAG of the proposed method, we denoted as Ours (only CAG) the success rate and computation time when CAG was used for the data whose planning failed in Ours w/o CAG.
Note that this success rate and computation time do not include the success rate and computation time when Ours w/o CAG succeeds.
These results are shown in Table~\ref{table:result_path}.
The success rate is the same as Section~\ref{sec:evaluate_mapping}, which is the percentage of trajectories that do not collide with any obstacles and satisfy equation~\eqref{eq:distance}.
Note that for the RRT and RRT Connect methods, if the execution time exceeded 60 s, the method was counted as a failure to find a path.

Ours w/o CAG is planning in the latent space where there is no collision with obstacles, so the computation time is almost constant, independent of the complexity of the environment.
However, learning-based methods alone cannot guarantee 100\% collision avoidance with obstacles.
In our results, the success rate of learning a trajectory without contact with obstacles was more than 60\%.
Even though the RRT and RRT Connect methods achieve a high success rate, these methods require more computation time as the complexity of the environment increases because the collision check is required each time.
Furthermore, the trajectory generated by each trial is different each time due to including randomness in RRT and RRT Connect.
In particular, when the complexity of the environment increases, the variance of generated trajectories becomes larger.
This experiment showed that the variance of the computation time and the length of the generated path was significant for the \emph{Upper-Bottom} and \emph{Bottom-Right} paths.
This means a complex planning task is included in the experiment.
Our method uses RRT Connect to compute the trajectory only before and after the collision.
Thus, the computation time is less than that of RRT or RRT Connect, which requires collision checks in all trajectories, as shown in Ours (only CAG).
The average computation time for Ours is $27.67\pm96.15~\mathrm{ms}$ in total, which is 21.8\% of that for RRT Connect of $126.79\pm44.84~\mathrm{ms}$.
While generating a single trajectory may result in collisions with obstacles in our method, generating multiple different trajectories, such as using different optimization criteria or selection of different trajectories in the latent space, can also reduce the likelihood of using RRT Connect.
Therefore, our proposed method requires even less computation than RRT and RRT Connect.

In summary, Ours and Ours w/o CAG require training time in advance, but the planning time is almost constant, independent of the complexity of the environment.
RRT and RRT Connect do not require learning, but the computation time increases with the complexity of the environment. Ours achieves both advantages of low-calculation cost with the learning method and guaranteed avoidance by the conventional planning method.
These indicate that the computational cost of our method is \textbf{\emph{scalable}} to the complexity of the environment.

\begin{figure}[t]
    \centering
    \includegraphics[width=0.95\columnwidth]{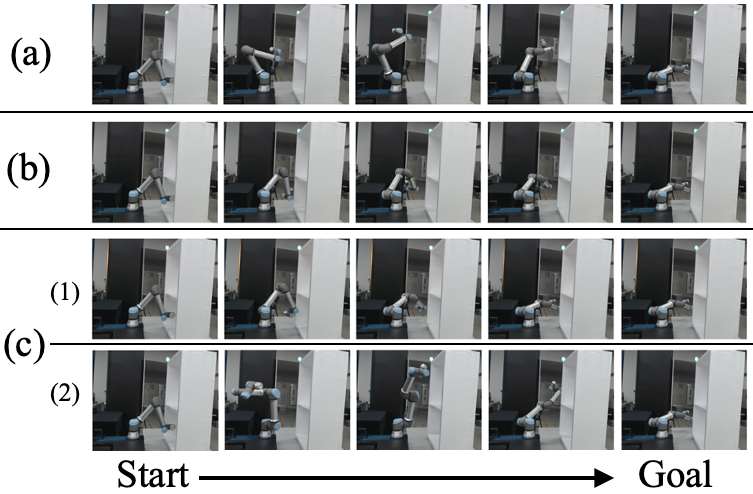}
    \caption{Demonstration on the actual robot. Different trajectories were generated for (a) to (c).
		(a) Ours without trajectory optimization, (b) Ours using trajectory optimization by speed minimization, (c) Generated trajectories using RRT Connect.}
    \label{fig:real_robot_demo}
\end{figure}

\subsection{Experiments Using UR5e}
\label{sec:result_ur5e}
In this section, we show the results of experiments using the actual UR5e.
Using the model trained with the simulation dataset, UR5e performed planning.
The shelf was placed in the same position as in the simulator, and the depth information was the same as in the simulator for the image.
In the experiments, we used (a) Ours without trajectory optimization, (b) Ours using trajectory optimization by speed minimization, and (c) Generated trajectories using RRT Connect.
Fig.~\ref{fig:real_robot_demo} shows the generated trajectories.
In (a), we confirm that our proposed method generates a collision-free path.
In (b), a shorter path is generated by optimizing for speed.
In (c), using RRT Connect, the trajectory generated by each trial is different each time.
On the other hand, in our method, if the same trajectory is selected in the latent space, the same trajectory will be generated in the joint space.

\section{Conclusion}
\label{sec:conclusion}
In this research, the robot's collision-free joint space is expressed as the latent space of cGANs, and collision-free paths are obtained by mapping the planning in the latent space to the joint space.
We confirmed that \rnum{1}~)~\textbf{\emph{Customizability}}; any path can be planned in the latent space using any optimization criteria,  \rnum{2}~)~\textbf{\emph{Adaptability}}; a single trained model could handle multiple untrained obstacle conditions, and \rnum{3}~)~\textbf{\emph{Scalability}}; computational cost of planning does not depend on the obstacles.
By modifying the trajectory in case of a collision when planning is done by learning alone, 100\% of collision avoidance can be guaranteed.
\section*{Appendix}
As a pre-experiment, we have published a paper in arXiv in which a 2-DoF robot arm is tested on a 2-D plane
\footnotemark{}.
Experiments are conducted in 2-D environments with random numbers and shapes of obstacles.
In the pre-experiment paper, the focus is on analysis because of the small number of DoFs.
In this new paper, the model is extended to a 6-DoF robot in 3-D space, but the concept of the model, which maps the non-collision posture to the potential space, is the same.
\footnotetext{The link of the article is the following: \url{https://arxiv.org/abs/2202.07203}}
\section*{Acknowledgments}
The authors would like to thank Avinash Ummadisingu for the proof check.
H. Mori would like to thank all colleagues in the ETIS lab at the Cergy-Pontoise Univ., especially Prof. Mathias Quoy, Prof. Philippe Gaussier, and Assoc. Prof. Alexandre Pitti for discussion about a preliminary result of the basic idea of this article when he came up with the basic idea at the lab in 2016.
\bibliographystyle{IEEEtran} 
\bibliography{IEEEabrv,bibliography} 
\end{document}